\newif\ifreport

\documentclass{article}
\reporttrue

\usepackage{mathtools}
\usepackage{tikz}
\usetikzlibrary{matrix,shapes,arrows,positioning,chains,patterns,fit,decorations.pathreplacing,calc,plotmarks}
\tikzset{
    dotted_block/.style={
        draw=black!30!white, 
        dashed,
        inner ysep=2mm,
        inner xsep=10mm, 
        rectangle, 
        rounded corners
    },
    block/.style={
        draw,
        rectangle,
        rounded corners,
        minimum height=2em,
        minimum width=2em
    },
    operator/.style={
        draw,
        circle,
        thin,
        minimum height=1em,
	   inner sep=1pt
    },
    weight/.style={
        draw,
        thin,
        rounded corners,
        rectangle,
    },
    value/.style={
        draw,
        thin,
        rectangle,
    },
    gain/.style={
        regular polygon, 
        regular polygon sides=3,
        draw, 
        fill=white, 
        text width=1em,
        inner sep=1mm, 
        outer sep=0mm,
        shape border rotate=-90
    },
    concat/.style={
        draw,
        shape=circle, 
        fill=black,
	   inner sep=0pt
    },
}

\usepackage{siunitx} 

\usepackage{longtable}

\usepackage{pgfplots}
\pgfplotsset{compat=1.17}
			 
\usepackage{booktabs}

\ifreport
	\title{Robust Recurrent Neural Network to Identify Ship Motion in Open Water with Performance Guarantees - Technical Report}
	\usepackage{jmlrutils}	
	\usepackage{PRIMEarxiv}
	\usepackage{hyperref}
	\usepackage{amsfonts}
	\usepackage{natbib}

	\newcommand{\acks}[1]{\section*{Acknowledgments}#1}
\else
	\title[Robust System Identification]{Robust Recurrent Neural Network to Identify Ship Motion in Open Water with Performance Guarantees}
\fi

\newtheorem{problem}[theorem]{Problem}

\usepackage{times}

\ifreport
	\author{
    	Daniel Frank, Decky Aspandi \\
    	Institute for Parallel and Distributed Systems \\
    	University of Stuttgart \\
    	\texttt{\{daniel.frank, decky.aspandi-latif\}@ipvs.uni-stuttgart} \\
	   \And
		Michael Muehlebach \\
		Max Planck Institute for Intelligent Systems \\
    	\texttt{michael.muehlebach@tuebingen.mpg.de} \\
    	\And
    	Benjamin Unger \\
    	Stuttgart Center for Simulation Science \\
        \texttt{benjamin.unger@simtech.uni-stuttgart.de} \\
 	      \And
		Steffen Staab \\
		Institute for Parallel and Distributed Systems \\
        University of Stuttgart and\\
		Electronics and Computer Science \\
		University of Southampton, UK \\
	   \texttt{steffen.staab@ipvs.uni-stuttgart} \\
 	}
\else
	\author{%
		\Name{Daniel Frank} \Email{daniel.frank@ipvs.uni-stuttgart.de} \\
		\Name{Decky Aspandi} \Email{Decky.Aspandi-Latif@ipvs.uni-stuttgart.de} \\
		\addr Institute for Parallel and Distributed Systems University of Stuttgart
		\AND
		\Name{Michael Muehlebach} \Email{michael.muehlebach@tuebingen.mpg.de} \\
		\addr Max Planck Institute for Intelligent Systems
		\AND
		\Name{Benjamin Unger} \Email{benjamin.unger@simtech.uni-stuttgart.de} \\
		\addr Stuttgart Center for Simulation Science
		\AND
		\Name{Steffen Staab} \Email{Steffen.Staab@ipvs.uni-stuttgart.de}\\
		\addr Institute for Parallel and Distributed Systems, University of Stuttgart and ECS, University of Southampton, UK
	}
\fi

\begin{document}

\maketitle

\begin{abstract}%
Recurrent neural networks are capable of learning the dynamics of an unknown nonlinear system purely from input-output measurements. However, the resulting models do not provide any stability guarantees on the input-output mapping. In this work, we represent a recurrent neural network as a linear time-invariant system with nonlinear disturbances. By introducing constraints on the parameters, we can guarantee finite gain stability and incremental finite gain stability. We apply this identification method to learn the motion of a four-degrees-of-freedom ship that is moving in open water and compare it against other purely learning-based approaches with unconstrained parameters. Our analysis shows that the constrained recurrent neural network has a lower prediction accuracy on the test set, but it achieves comparable results on an out-of-distribution set and respects stability conditions.
\end{abstract}

\ifreport
\fi

\section{Introduction}
Traditional system identification often relies on domain-specific knowledge to obtain a representation of the target system. Thereby, complex or unknown effects are often neglected or simplified, which limits the accuracy of the mathematical model. In contrast, purely learning-based approaches have been shown to predict system states of an unknown nonlinear system with high accuracy. Due to a large number of parameters, it is assumed that deep neural networks can learn the internal dynamics, which are not directly visible in the input-output measurements. However, compared to models relying on differential equations, which are derived using traditional modeling techniques, it is much more difficult to satisfy stability requirements on the identified system.
Recurrent neural networks~(RNN) can handle input sequences and predict future system states of unknown systems, but they generally lack stability guarantees. From a robust control perspective, RNNs can be modeled as linear time-invariant systems with nonlinear disturbances. This perspective allows us to formulate constraints on the learnable parameters that guarantee finite gain and incremental finite gain stability. This is achieved by bounding the nonlinear activation function of the RNN with (incremental) quadratic constraints.
We compare the constrained RNN with purely learning-based models such as long short-term memory (LSTM) \citep{hochreiter1997long} and unconstrained RNN \citep{goodfellow2016deep} that both achieve high prediction accuracy when learning the ship motion. We evaluate the different systems on within-distribution and out-of-distribution (OOD) test data. Our experiment results indicate that RNNs with parametric constraints do not achieve the same level of accuracy on within-distribution data as purely learning-based models without parametric constrains. However, they guarantee upper bounds on the finite (incremental) stability gain and are on par with an OOD evaluation.

Thus, our main contributions can be summarized as follows: (i) We apply a robust RNN model that guarantees a finite (incremental) stability gain to identify a real-world system (ship motion in open water) with multiple inputs and multiple outputs. (ii) We provide a comprehensive overview of robust RNNs, which is rooted in control theory and leads to guaranteed finite (incremental) stability gains. (iii) We evaluate robust RNNs and comparable learning-based approaches on ship motion data, including OOD measurements and empirically analyze the stability gains to support our theoretical results.

The article is structured as follows: In \sectionref{sec:related_work}, we review existing approaches, whereas \sectionref{sec:background} introduces the theoretical background and the notation that is subsequently used. In \sectionref{sec:methodology}, we introduce our robust RNN model and explain how it is trained. The model is evaluated and compared to other approaches from the literature in \sectionref{sec:numerical_experiment}. The article concludes in \sectionref{sec:conclusion}.

\section{Related Work}\label{sec:related_work}
Deep recurrent neural network structures showed high prediction accuracy on nonlinear system identification tasks such as quadrotor- \citep{mohajerin2019multistep} or ship-motion \citep{baier2021hybrid}. Compared to classical identification approaches, which are extensively studied in \citet{pillonetto2022classical}, these learning methods usually do not provide stability guarantees on the input-output behavior.
Separating the nonlinear activation function from the linear layers in a neural network allows us to use classical tools from robust control to analyze stability, which was already proposed by \citet{suykens1995nonlinear}. More recently, \citet{fazlyab2019efficient} used semi-definite programming to calculate a Lipschitz gain for deep neural networks. For sequence to sequence models that are used in system identification, \citet{revay2020convex} introduced convex parametric constraints to guarantee finite incremental gain stability for recurrent neural networks. In \citet{revay2021recurrent}, this network structure is generalized to equilibrium networks, which do not require parametric constraints. In \citep{junnarkar2022synthesis}, a recurrent neural network is used to control a partially unknown linear system, whereby closed-loop stability guarantees are established. An extension to recurrent equilibrium models is provided in \citet{gu2022recurrent}. The incremental stability certificates achieved by the aforementioned approaches rely on a contraction argument. In contrast, \citet{pauli2022robustness} establishes finite gain stability by dissipativity arguments, which also allows for other performance specifications.
In this work, we follow the approach of \citet{revay2020convex} and show finite gain stability using arguments from \citet{pauli2022robustness}. Compared to existing methods, we are evaluating the identification model on a real-world problem with multiple input and multiple outputs and compare against state of the art learning-based approaches \citep{mohajerin2019multistep}.

\section{Background}\label{sec:background}
\subsection*{Notation}
With $u = (u^k)_{k=0}^\infty$ we refer to a sequence of vectors. We distinguish two sequences $u_a, u_b$ by sub-indices $a,b$. The $k$-th vector in the sequence $u$ is denoted by $u^k\in \mathbb{R}^{n_u}$, where $k$ denotes the discrete time index $k\in [0,\ldots,T]$. The variables $u^k_m$ and $u^k_{m,a}$ refer to the $m$-th component of the vectors $u^k$ and $u^k_{\cdot,a}$, respectively. If no sub- or superscript is present, we refer to the entire sequence. In this work, we consider discrete-time systems and square-summable sequences which are from the set $\ell_{2e}^{n_u}\vcentcolon=\{(u^k)_{k=0}^{\infty} \mid u^k \in \mathbb{R}^{n_u}, ~\sum_{k=0}^T \|u^k\|^2 < \infty \text{ for all } T>0\}$ with squared Euclidean norm $\|u^k\|^2 \vcentcolon= \sum_{i=1}^{n_u} (u_i^k)^2 = (u^k)^{\mathrm{T}} (u^k) $. We refer to $\sum_{k=0}^T\|\cdot\|^2$ as the squared $\ell_2$-norm and use the notation $\Delta u^k = u^k_a - u^k_b$ and $\Delta \psi(z^k) = \psi(z^k_a)- \psi(z^k_b)$ to denote the difference at time step $k$. With $I$, we denote the identity matrix of appropriate size. We use $M\succeq 0$ ($M\preceq 0$), to indicate that $M$ is a positive semi-definite (negative semi definite) and $M\succ 0$ ($M\prec 0$), to indicate that $M$ is a positive definite (negative definite). For matrices that can be inferred from symmetry, we use the symbol $\star$.

\subsection*{Model Structure and Robust Performance}
The input-output behavior of a dynamical system can be described in various ways and is often useful for characterizing the performance of a system (a more precise definition can be found in \citet[Def.\,4]{veenman2016robust}. Our main result that guarantees a finite (incremental) stability gain for RNN is one specific performance measure of a more general framework used in robust control. We refer to \citet{scherer2006lmi} for more details and the discussion of other performance measures. For system identification tasks, we focus on the worst possible amplification and the sensitivity, which refers to the finite stability gain and the finite incremental stability gain respectively, both performance measures will be defined in this subsection.

First, we introduce the model structure. We consider the system $\mathcal{S}$ as a linear system $G$ in feedback interconnection with a nonlinearity $\psi$. The interconnection is shown in \figureref{fig:feedback_inter} and formalized as:
\begin{equation}\label{eq:system_s}
    G \colon 
    \left\{ \begin{aligned}
        x^{k+1} & = Ax^k + B_1 u^k + B_2 w^k \\
        \hat{y}^k & = C_1x^k + D_{11} u^k + D_{12} w^k \\
        z^k & = C_2 x^k + D_{21} u^k
        \end{aligned}\right. , \qquad 
        w^k = \psi(z^k).
\end{equation}
The system $\mathcal{S}$ maps an input sequence $u \in \ell_{2e}^{n_u}$ to an output sequence $\hat{y} = S(u) \in \ell_{2e}^{n_y}$.

Next, we review the necessary performance conditions that our models should satisfy.
\begin{definition}[Finite gain stability]\label{def:finite_gain_stability}
    \citep[cf.][Def.\,4.14]{sastry2013nonlinear} The system $\mathcal{S}$ is finite gain $\ell_2$ stable if there exists $\gamma, \gamma_0 > 0$ such that for a given $u\in \ell_{2e}^{n_u}$ the output $\hat{y} = \mathcal{S}(u)\in \ell_{2e}^{n_y}$ satisfies
    \begin{equation}
        \sum_{k=0}^T\|\hat{y}^k\|^2 \leq \gamma^2 \sum_{k=0}^T\|u^k \|^2 + \gamma_0 \qquad \text{for all } T>0.
        \label{eq:l2_gain}
    \end{equation}
\end{definition}
\begin{definition}[Incremental finite gain stability]\label{def:icnremental_stability}
    \citep[cf.][Def.\,4.16]{sastry2013nonlinear} The system $\mathcal{S}$ is said to be incrementally finite gain stable if there exists $\gamma_{\mathrm{inc}}>0$ such that for all $u_a, u_b \in \ell_{2e}^{n_u}$
    \begin{equation}
        \sum_{k=0}^T\| \Delta \hat{y}^k\|^2 \leq \gamma_{\mathrm{inc}}^2 \sum_{k=0}^T\|\Delta u^k \|^2 \qquad \text{for all } T>0,
        \label{eq:incremental_l_2_gain}
    \end{equation}  
    holds, where $\Delta u = u_a - u_b$ and $\Delta \hat{y} = \mathcal{S}(u_a) - \mathcal{S}(u_b)$.
\end{definition}
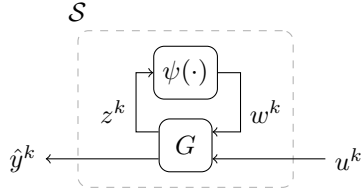
\begin{figure}
    \centering
    \begin{tikzpicture}[node distance = 0.25cm and 0.7cm, auto, align=center]

    \node[] (output) {};
    \node[block, right= of output] (G) {$G$};
    \node[block, above= of G] (delta) {$\psi(\cdot)$};
    \node[right= of G] (input) {};
    \node[dotted_block, fit = (G) (delta)] (S) {};
    \node at (S.north west) [above] {$\mathcal{S}$};

    \coordinate[] (outputz)  at ($(G.south west)!0.75!(G.north west)$);
    \coordinate[] (outputy)  at ($(G.south west)!0.25!(G.north west)$);
    \coordinate[] (inputw) at ($(G.south east)!0.75!(G.north east)$);
    \coordinate[] (inputu) at ($(G.south east)!0.25!(G.north east)$);
    
    \draw[<-] (inputu) -- ++(1.5,0);
    \draw[-] (inputu) ++(1.5,0) node[right]{$u^k$}++(0.5,0);
    \draw[->] (outputz) -- ++(-0.3,0) node[above left]{$z^k$}  |-   (delta.west) ;
    \draw[->] (outputy)  --  ++(-1.5,0);
    \draw[-] (outputy)  ++(-1.5,0) node[left]{$\hat{y}^k$} ++(-1,0);
    \draw[->] (delta.east)  -- ++(0.3,0) |- node[above right] {$w^k$}(inputw) ;

\end{tikzpicture}
    \caption{Interconnected system $\mathcal{S}$, consists of a linear system $G$ and nonlinear disturbance $\psi(\cdot)$.}
    \label{fig:feedback_inter}
\end{figure}

\begin{remark}
    Incremental finite gain stability implies finite gain stability. This follows directly from Definition~\ref{def:icnremental_stability} if one input sequence (e.g.~$u_b$) is set to zero.
\end{remark}
To ensure robust performance, we use the following result from \citet{scherer2006lmi}.
\begin{theorem}[Robust Quadratic Performance]\label{thm:robust_quadratic_performance}
    The interconnected system $\mathcal{S}$ is well-posed and satisfies robust quadratic performance if there exists $X \succ 0$ and
    \begin{equation}\label{eq:uncertainty_condition}
        P=\begin{pmatrix}Q & S \\S^{\mathrm{T}} & R\end{pmatrix} \qquad \text{ with } \qquad
        \begin{pmatrix}
            \psi(z^k) \\
            z^k
        \end{pmatrix}^{\mathrm{T}} P
        \begin{pmatrix}
            \psi(z^k) \\
            z^k
        \end{pmatrix} \geq 0 \qquad \text{for all } z^k \in \ell^{n_z}_{2e}
    \end{equation}
    and performance matrices $Q_p$, $R_p$, $S_p$ such that
    \begin{equation}\label{eq:robust_performance_condition}
        \begin{split}
            \begin{pmatrix}
               \star
            \end{pmatrix}^{\mathrm{T}}
            \begin{pmatrix}
                -X & 0\\
                0 & X
            \end{pmatrix}
            \begin{pmatrix}
                I & 0 & 0 \\
                A & B_1 & B_2
            \end{pmatrix}+
            \begin{pmatrix}
                \star
            \end{pmatrix}^{\mathrm{T}}
            \begin{pmatrix}
                Q_p & S_p \\
                S_p^{\mathrm{T}} & R_p
            \end{pmatrix}
            \begin{pmatrix}
                0 & I & 0\\
                C_1 & D_{11} & D_{12}
            \end{pmatrix}+& \\
            \begin{pmatrix}
               \star
            \end{pmatrix}^{\mathrm{T}}
            P
            \begin{pmatrix}
                0 & 0 & I \\
                C_2 & D_{12} & 0
            \end{pmatrix} & \prec 0.
        \end{split}
    \end{equation}
\end{theorem}
We sketch the key ideas of the proof and refer to \citet{scherer2006lmi} or more recently \citet[Thm. 1]{pauli2022robustness}, for more details \ifreport (cf. proof of \corollaryref{cor:l2_bounded}) \fi. Well-posedness is given for $R_p\succeq 0$ and condition \eqref{eq:uncertainty_condition}. It can be shown that \eqref{eq:l2_gain} holds by left multiplying \eqref{eq:robust_performance_condition} with $\begin{bmatrix}(x^k)^{\mathrm{T}} & (u^k)^{\mathrm{T}} & (w^k)^{\mathrm{T}}\end{bmatrix}$ and right multiplying its transpose. The key point is to cancel the term \eqref{eq:uncertainty_condition} in the resulting inequality.

\section{Methodology}\label{sec:methodology}
\subsection{Robust System Identification with Linear Matrix Inequality Constraints}
In this section, we develop parametric constraints for $G$ such that the stability properties from the previous section are satisfied. We further explain how these constraints are satisfied during training and how the initial state $x^0$ and the initial parameter set $\theta^0$ is chosen.
\begin{remark}
    We follow the same ideas as presented in \citet{revay2020convex} and extend the results to guarantee finite gain stability.
\end{remark}
Our goal is to learn the dynamics of a nonlinear system purely from data, while satisfying stability properties according to Definition \ref{def:finite_gain_stability} and \ref{def:icnremental_stability}. This is summarized in the following problem statement.
\begin{problem}\label{pb:learn_linear_system_parameters}
    Given a dataset $\mathcal{D} = \left\{ (u, y)_i\right\}$ for $i=1, \ldots, K$ where $u\in\ell_{2e}^{n_u}$ and $y\in\ell_{2e}^{n_y}$ are measurements taken from an unknown dynamical system, find the parameters $$\theta = \left\{A,~B_1,~B_2,~C_1,~D_{11}, ~D_{12}, ~C_2, ~D_{21} \right\}$$ of $G$ that minimize the loss $\sum_{k=0}^T\|\hat{y}^k - y^k\|^2$ and satisfy \eqref{eq:l2_gain} and \eqref{eq:incremental_l_2_gain} for some $\gamma^2$ and $\gamma_{\mathrm{inc}}^2$.
\end{problem}
To apply \theoremref{thm:robust_quadratic_performance}, we bound the nonlinearity by linear functions. Nonlinear activation functions commonly used in deep learning like $\operatorname{ReLU}(\cdot)$ or $\tanh(\cdot)$ are sector bounded \citep[e.g.~used in][]{pauli2022robustness}
\begin{equation}
    (\alpha a - \varphi(a))(\varphi(a) - \beta a) \geq 0 \qquad \forall a \in \mathbb{R}\label{eq:sector_bounded}
\end{equation} and slope restricted \citep[e.g.~used in][]{fazlyab2019efficient}
\begin{equation}
    \begin{aligned}
    (\mu \Delta_{a,b} - \Delta_{a,b}\varphi)(\Delta_{a,b}\varphi - \eta \Delta_{a,b}) \geq 0,\\\Delta_{a,b} \vcentcolon= a-b,~\Delta_{a,b}\varphi\vcentcolon=\varphi(a) -\varphi(b),
    \end{aligned}\qquad \forall a,b\in \mathbb{R},~a\neq b. \label{eq:slope_restriction}
\end{equation}
For instance, $\operatorname{ReLU}(\cdot)$ or $\tanh(\cdot)$ satisfies \eqref{eq:sector_bounded} with $\alpha=0$, $\beta=1$ and \eqref{eq:slope_restriction} with $\mu=0$, $\eta=1$. In this work, we focus on $ \varphi = \tanh$ as it is often used for RNNs \citep{goodfellow2016deep}. 
\ifreport
    The sector and slope condition is visualized in \figureref{fig:bounded_nonlinearity}.
    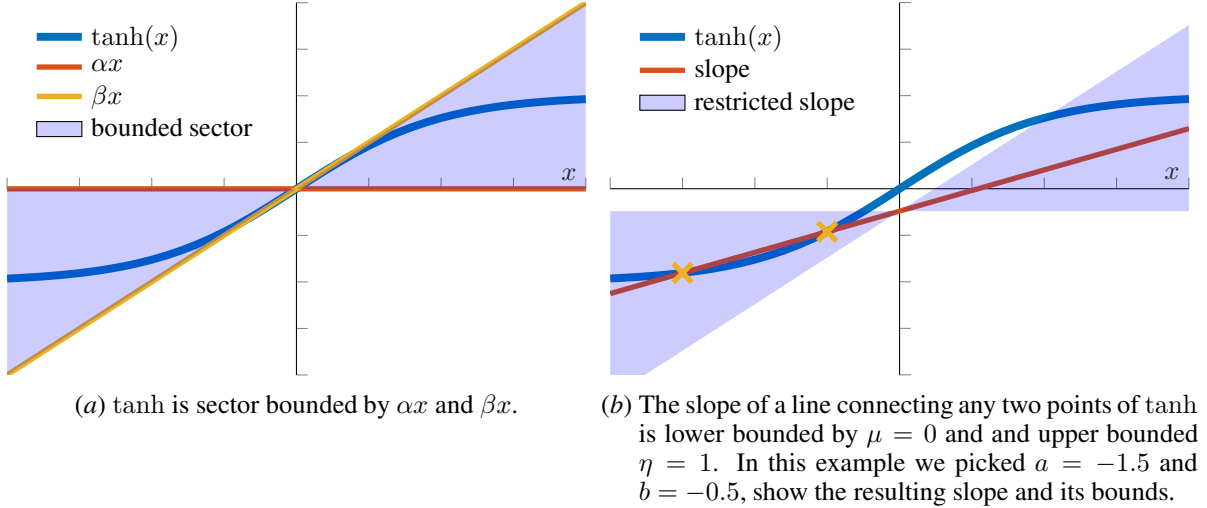
\begin{figure}
        \floatconts
        {fig:bounded_nonlinearity}
        {\caption{Activation function $\tanh$ is sector bounded \eqref{eq:sector_bounded} and slope restricted \eqref{eq:slope_restriction}}}
        {%
        \subfigure[$\tanh$ is sector bounded by $\alpha x$ and $\beta x$.]{\label{fig:sector_bound}%
%
\definecolor{mycolor1}{rgb}{0.00000,0.44700,0.74100}%
\definecolor{mycolor2}{rgb}{0.85000,0.32500,0.09800}%
\definecolor{mycolor3}{rgb}{0.92900,0.69400,0.12500}%
\begin{tikzpicture}

\begin{axis}[%
width=0.466\textwidth,
height=0.3\textwidth,
at={(0\textwidth,0\textwidth)},
scale only axis,
xmin=-2,
xmax=2,
xtick={-2,-1.5,-1,-0.5,0,0.5,1,1.5,2},
xticklabels={\empty},
xlabel style={font=\color{white!15!black}, at={(ticklabel* cs:1)}, anchor=south east},
xlabel={$x$},
ymin=-2,
ymax=2,
ytick={-2,-1.5,-1,-0.5,0,0.5,1,1.5,2},
yticklabels={\empty},
axis background/.style={fill=white},
axis x line*=middle,
axis y line*=middle,
legend style={at={(0.03,0.97)}, anchor=north west, legend cell align=left, align=left, fill=none, draw=none}
]
\addplot [color=mycolor1, line width=3.0pt]
  table[row sep=crcr]{%
-2	-0.964027580075817\\
-1.9	-0.956237458127739\\
-1.8	-0.946806012846268\\
-1.7	-0.935409070603099\\
-1.6	-0.921668554406471\\
-1.5	-0.905148253644866\\
-1.4	-0.885351648202263\\
-1.3	-0.861723159313306\\
-1.2	-0.833654607012155\\
-1.1	-0.80049902176063\\
-1	-0.761594155955765\\
-0.9	-0.716297870199024\\
-0.8	-0.664036770267849\\
-0.7	-0.604367777117164\\
-0.6	-0.537049566998035\\
-0.5	-0.46211715726001\\
-0.4	-0.379948962255225\\
-0.3	-0.291312612451591\\
-0.2	-0.197375320224904\\
-0.0999999999999999	-0.0996679946249557\\
0	0\\
0.0999999999999999	0.0996679946249557\\
0.2	0.197375320224904\\
0.3	0.291312612451591\\
0.4	0.379948962255225\\
0.5	0.46211715726001\\
0.6	0.537049566998035\\
0.7	0.604367777117164\\
0.8	0.664036770267849\\
0.9	0.716297870199024\\
1	0.761594155955765\\
1.1	0.80049902176063\\
1.2	0.833654607012155\\
1.3	0.861723159313306\\
1.4	0.885351648202263\\
1.5	0.905148253644866\\
1.6	0.921668554406471\\
1.7	0.935409070603099\\
1.8	0.946806012846268\\
1.9	0.956237458127739\\
2	0.964027580075817\\
};
\addlegendentry{$\tanh(x)$}

\addplot [color=mycolor2, line width=2.0pt]
  table[row sep=crcr]{%
-2	-0\\
-1.9	-0\\
-1.8	-0\\
-1.7	-0\\
-1.6	-0\\
-1.5	-0\\
-1.4	-0\\
-1.3	-0\\
-1.2	-0\\
-1.1	-0\\
-1	-0\\
-0.9	-0\\
-0.8	-0\\
-0.7	-0\\
-0.6	-0\\
-0.5	-0\\
-0.4	-0\\
-0.3	-0\\
-0.2	-0\\
-0.0999999999999999	-0\\
0	0\\
0.0999999999999999	0\\
0.2	0\\
0.3	0\\
0.4	0\\
0.5	0\\
0.6	0\\
0.7	0\\
0.8	0\\
0.9	0\\
1	0\\
1.1	0\\
1.2	0\\
1.3	0\\
1.4	0\\
1.5	0\\
1.6	0\\
1.7	0\\
1.8	0\\
1.9	0\\
2	0\\
};
\addlegendentry{$\alpha x$}

\addplot [color=mycolor3, line width=2.0pt]
  table[row sep=crcr]{%
-2	-2\\
-1.9	-1.9\\
-1.8	-1.8\\
-1.7	-1.7\\
-1.6	-1.6\\
-1.5	-1.5\\
-1.4	-1.4\\
-1.3	-1.3\\
-1.2	-1.2\\
-1.1	-1.1\\
-1	-1\\
-0.9	-0.9\\
-0.8	-0.8\\
-0.7	-0.7\\
-0.6	-0.6\\
-0.5	-0.5\\
-0.4	-0.4\\
-0.3	-0.3\\
-0.2	-0.2\\
-0.0999999999999999	-0.0999999999999999\\
0	0\\
0.0999999999999999	0.0999999999999999\\
0.2	0.2\\
0.3	0.3\\
0.4	0.4\\
0.5	0.5\\
0.6	0.6\\
0.7	0.7\\
0.8	0.8\\
0.9	0.9\\
1	1\\
1.1	1.1\\
1.2	1.2\\
1.3	1.3\\
1.4	1.4\\
1.5	1.5\\
1.6	1.6\\
1.7	1.7\\
1.8	1.8\\
1.9	1.9\\
2	2\\
};
\addlegendentry{$\beta x$}

\addplot[area legend, draw=none, fill=blue, fill opacity=0.2]
table[row sep=crcr] {%
x	y\\
0	0\\
-2	0\\
-2	-2\\
}--cycle;
\addlegendentry{bounded sector}

\addplot[area legend, draw=none, fill=blue, fill opacity=0.2, forget plot]
table[row sep=crcr] {%
x	y\\
0	0\\
2	0\\
2	2\\
}--cycle;
\end{axis}
\end{tikzpicture}%
}
        \subfigure[The slope of a line connecting any two points of $\tanh$ is lower bounded by $\mu=0$ and and upper bounded $\eta=1$. In this example we picked $a = -1.5$ and $b=-0.5$, show the resulting slope and its bounds.]{\label{fig:slope_restricted}%
%
\definecolor{mycolor1}{rgb}{0.00000,0.44700,0.74100}%
\definecolor{mycolor2}{rgb}{0.85000,0.32500,0.09800}%
\definecolor{mycolor3}{rgb}{0.92900,0.69400,0.12500}%
\begin{tikzpicture}

\begin{axis}[%
width=0.466\textwidth,
height=0.3\textwidth,
at={(0\textwidth,0\textwidth)},
scale only axis,
xmin=-2,
xmax=2,
xtick={-2,-1.5,-1,-0.5,0,0.5,1,1.5,2},
xticklabels={\empty},
xlabel style={font=\color{white!15!black}, at={(ticklabel* cs:1)}, anchor=south east},
xlabel={$x$},
ymin=-2,
ymax=2,
ytick={-2,-1.5,-1,-0.5,0,0.5,1,1.5,2},
yticklabels={\empty},
axis background/.style={fill=white},
axis x line*=middle,
axis y line*=middle,
legend style={at={(0.03,0.97)}, anchor=north west, legend cell align=left, align=left, fill=none, draw=none}
]
\addplot [color=mycolor1, line width=3.0pt]
  table[row sep=crcr]{%
-2	-0.964027580075817\\
-1.9	-0.956237458127739\\
-1.8	-0.946806012846268\\
-1.7	-0.935409070603099\\
-1.6	-0.921668554406471\\
-1.5	-0.905148253644866\\
-1.4	-0.885351648202263\\
-1.3	-0.861723159313306\\
-1.2	-0.833654607012155\\
-1.1	-0.80049902176063\\
-1	-0.761594155955765\\
-0.9	-0.716297870199024\\
-0.8	-0.664036770267849\\
-0.7	-0.604367777117164\\
-0.6	-0.537049566998035\\
-0.5	-0.46211715726001\\
-0.4	-0.379948962255225\\
-0.3	-0.291312612451591\\
-0.2	-0.197375320224904\\
-0.0999999999999999	-0.0996679946249557\\
0	0\\
0.0999999999999999	0.0996679946249557\\
0.2	0.197375320224904\\
0.3	0.291312612451591\\
0.4	0.379948962255225\\
0.5	0.46211715726001\\
0.6	0.537049566998035\\
0.7	0.604367777117164\\
0.8	0.664036770267849\\
0.9	0.716297870199024\\
1	0.761594155955765\\
1.1	0.80049902176063\\
1.2	0.833654607012155\\
1.3	0.861723159313306\\
1.4	0.885351648202263\\
1.5	0.905148253644866\\
1.6	0.921668554406471\\
1.7	0.935409070603099\\
1.8	0.946806012846268\\
1.9	0.956237458127739\\
2	0.964027580075817\\
};
\addlegendentry{$\tanh(x)$}

\addplot [color=mycolor2, line width=2.0pt]
  table[row sep=crcr]{%
-2	-1.12666380183729\\
-1.9	-1.08236069219881\\
-1.8	-1.03805758256032\\
-1.7	-0.993754472921838\\
-1.6	-0.949451363283352\\
-1.5	-0.905148253644866\\
-1.4	-0.860845144006381\\
-1.3	-0.816542034367895\\
-1.2	-0.772238924729409\\
-1.1	-0.727935815090924\\
-1	-0.683632705452438\\
-0.9	-0.639329595813952\\
-0.8	-0.595026486175467\\
-0.7	-0.550723376536981\\
-0.6	-0.506420266898495\\
-0.5	-0.46211715726001\\
-0.4	-0.417814047621524\\
-0.3	-0.373510937983038\\
-0.2	-0.329207828344553\\
-0.0999999999999999	-0.284904718706067\\
0	-0.240601609067581\\
0.0999999999999999	-0.196298499429096\\
0.2	-0.15199538979061\\
0.3	-0.107692280152125\\
0.4	-0.0633891705136388\\
0.5	-0.0190860608751531\\
0.6	0.0252170487633325\\
0.7	0.0695201584018182\\
0.8	0.113823268040304\\
0.9	0.15812637767879\\
1	0.202429487317275\\
1.1	0.246732596955761\\
1.2	0.291035706594246\\
1.3	0.335338816232732\\
1.4	0.379641925871218\\
1.5	0.423945035509704\\
1.6	0.468248145148189\\
1.7	0.512551254786675\\
1.8	0.556854364425161\\
1.9	0.601157474063646\\
2	0.645460583702132\\
};
\addlegendentry{slope}

\addplot[area legend, draw=none, fill=blue, fill opacity=0.2]
table[row sep=crcr] {%
x	y\\
0	-0.240601609067581\\
-2	-0.240601609067581\\
-2	-2.24060160906758\\
}--cycle;
\addlegendentry{restricted slope}

\addplot[area legend, draw=none, fill=blue, fill opacity=0.2, forget plot]
table[row sep=crcr] {%
x	y\\
0	-0.240601609067581\\
2	-0.240601609067581\\
2	1.75939839093242\\
}--cycle;
\addplot [color=mycolor3, line width=2.0pt, only marks, mark size=5.0pt, mark=x, mark options={solid, mycolor3}, forget plot]
  table[row sep=crcr]{%
-1.5	-0.905148253644866\\
-0.5	-0.46211715726001\\
};
\end{axis}
\end{tikzpicture}%
}
        }
    \end{figure}
\fi
Since the activation function $\varphi\colon\mathbb{R} \to \mathbb{R}$ is applied to every neuron in the hidden layer, we define $\psi(z^k) \vcentcolon= \left[ \varphi(z_1^k), \ldots, \varphi(z_{n_z}^k)\right]^{\mathrm{T}}$, which is the disturbance block in \figureref{fig:feedback_inter}. Next, we introduce static diagonal multipliers $T = \operatorname{diag}(\lambda_1, \ldots, \lambda_{n_z})$, $\lambda_i >0$, and reformulate condition \eqref{eq:sector_bounded} as a quadratic constraint and condition \eqref{eq:slope_restriction} as an incremental quadratic constraint including multipliers as follows:
\begin{equation}\label{eq:P_lmi}
    \begin{pmatrix}
        \star
    \end{pmatrix}^{\mathrm{T}}
    \underbrace{\begin{pmatrix}
        -2T & (\alpha+\beta)T \\
        (\alpha+\beta)T & -2\alpha\beta T
    \end{pmatrix}}_{=: P_{\text{sect}}}
    \begin{pmatrix}
        \psi( z^k ) \\
        z^k
    \end{pmatrix} \geq 0, 
    \begin{pmatrix}
        \star
    \end{pmatrix}^{\mathrm{T}}
    \underbrace{
        \begin{pmatrix}
            -2T & (\mu+\eta)T \\
            (\mu+\eta)T & -2\mu\eta T
        \end{pmatrix}
    }_{=: P_{\text{slope}}}
    \begin{pmatrix}
        \Delta \psi( z^k ) \\
        \Delta z^k
    \end{pmatrix} \geq 0. 
\end{equation}
\theoremref{thm:robust_quadratic_performance} is classically used to verify robust quadratic performance of a known linear system~$G$. However, for system identification problems, the parameters $\theta$ are unknown. Condition \eqref{eq:robust_performance_condition} is not convex with respect to the parameter set $\theta$ and therefore cannot be solved by semi-definite programming. We refer to \citet{boyd2004convex} for the topic of convex optimization and semi-definite programming. To convexify condition \eqref{eq:robust_performance_condition}, we introduce new parameters $\tilde{\theta}\vcentcolon= \left\{\tilde{A}, \tilde{B}_1, \tilde{B}_2, C_1, D_{11}, D_{12}, \tilde{C}_2, \tilde{D}_{21}\right\}$, with $\tilde{A} \vcentcolon= XA$, $\tilde{B}_1 \vcentcolon= XB_1$, $\tilde{B}_2 \vcentcolon= XB_2$, $\tilde{C}_2 \vcentcolon= TC_2$, $\tilde{D}_{21} \vcentcolon= TD_{21}$, which yields the linear matrix inequality 
\begin{equation}
        M \vcentcolon= \begin{pmatrix}
            -X & 0 & \tilde{C}_2^{\mathrm{T}} & \tilde{A}^{\mathrm{T}} & C_1^{\mathrm{T}} \\
            0 & -\boldsymbol{\gamma}^2 I & \tilde{D}_{21}^{\mathrm{T}} & \tilde{B}_1^{\mathrm{T}}  & D_{11}^{\mathrm{T}} \\
            \tilde{C}_2 & \tilde{D}_{21} & -2T & \tilde{B}_2^{\mathrm{T}} & D_{12}^{\mathrm{T}} \\
            \tilde{A} & \tilde{B}_1 & \tilde{B}_2  & -X & 0 \\
            C_1 & D_{11} & D_{12} & 0 & -I
        \end{pmatrix}\prec 0,
    \label{eq:linear_constr_lmi}
\end{equation}
where $\boldsymbol{\gamma} \in \left\{\gamma,~\gamma_{\text{inc}}\right\}$. With the constraints \eqref{eq:P_lmi} and \eqref{eq:linear_constr_lmi}, we can now state the main stability results.
\begin{corollary}
    \label{cor:l2_bounded}
    If the parameter set $\tilde{\theta}$ satisfies constraint \eqref{eq:linear_constr_lmi}, the system $\mathcal{S}$ defined in \eqref{eq:system_s} has a finite $\ell_2$-gain $\gamma^2$ and satisfies stability condition \eqref{eq:l2_gain}.
\end{corollary}

\ifreport
    \begin{proof}
        We follow standard arguments of \citet{scherer2000linear} to show that \eqref{eq:l2_gain} holds for the interconnection $\mathcal{S}$, that was recently also used in \citep{pauli2022robustness}.

        Firstly, we use the Schur complement to obtain:
        $$
        \begin{pmatrix}
            \tilde{A}^{\mathrm{T}}  & C_1^{\mathrm{T}}\\
            \tilde{B}_1^{\mathrm{T}} & D_{11}^{\mathrm{T}}\\
            \tilde{B}_2^{\mathrm{T}} & D_{12}^{\mathrm{T}}
        \end{pmatrix}
        \begin{pmatrix}
            X^{-1} & 0 \\
            0 & I
        \end{pmatrix}
        \begin{pmatrix}
            \tilde{A} & \tilde{B}_1 & \tilde{B}_2 \\
            C_1 & D_{11} & D_{12}
        \end{pmatrix} +
        \begin{pmatrix}
            -X & 0 & \tilde{C}_2^{\mathrm{T}} \\
            0 & -\gamma^2 I & \tilde{D}_{21}^{\mathrm{T}} \\
            \tilde{C}_2 & \tilde{D}_{21} & -2I
        \end{pmatrix} \prec 0
        $$
        Since \eqref{eq:linear_constr_lmi} is satisfied $X$, $T$ are invertible and $X = X^{\mathrm{T}} X^{-1}X$ holds, we can insert the original parameters of \eqref{eq:system_s} to get the structure from \theoremref{thm:robust_quadratic_performance}
        \begin{equation}\label{eq:robust_performance_condition_proof}
            \begin{split}
                \begin{pmatrix}
                    I & 0 & 0 \\
                    A & B_1 & B_2
                \end{pmatrix}^{\mathrm{T}}
                \begin{pmatrix}
                    -X & 0\\
                    0 & X
                \end{pmatrix}
                \begin{pmatrix}
                    I & 0 & 0 \\
                    A & B_1 & B_2
                \end{pmatrix}+ \\
                \begin{pmatrix}
                    0 & I & 0\\
                    C_1 & D_{11} & D_{12}
                \end{pmatrix}^{\mathrm{T}}
                \begin{pmatrix}
                    -\gamma I & 0\\
                    0 & I
                \end{pmatrix}
                \begin{pmatrix}
                    0 & I & 0\\
                    C_1 & D_{11} & D_{12}
                \end{pmatrix}+ \\
                \begin{pmatrix}
                    0 & 0 & I \\
                    C_2 & D_{12} & 0
                \end{pmatrix}^{\mathrm{T}}
                P_{\text{sect}}
                \begin{pmatrix}
                    0 & 0 & I \\
                    C_2 & D_{12} & 0
                \end{pmatrix} \prec 0.
            \end{split}
        \end{equation}
        The variable $z^k$ does not depend on $w^k$, therefore well-posedness of the interconnection directly follows from \eqref{eq:P_lmi}. We left multiply \eqref{eq:robust_performance_condition} by $\begin{bmatrix}(x^k)^{\mathrm{T}} & (u^k)^{\mathrm{T}} & (w^k)^{\mathrm{T}}\end{bmatrix}$ and right multiply its transpose to arrive at
        \begin{equation} \label{eq:dissipativity_0}
            \begin{pmatrix}
                x^{k+1} \\
                x^k
            \end{pmatrix}^{\mathrm{T}}
            \begin{pmatrix}
                -X & 0 \\
                0 & X
            \end{pmatrix}
            \begin{pmatrix}
                x^{k+1} \\
                x^k
            \end{pmatrix}+
            \begin{pmatrix}
                u^{k} \\
                y^k
            \end{pmatrix}^{\mathrm{T}}
            \begin{pmatrix}
                -\gamma^2 I & 0 \\
                0 & I
            \end{pmatrix}
            \begin{pmatrix}
                u^{k} \\
                y^k
            \end{pmatrix}+
            \begin{pmatrix}
                w^k \\
                z^k
            \end{pmatrix}^{\mathrm{T}}
            P_{\text{sect}}
            \begin{pmatrix}
                w^k \\
                z^k
            \end{pmatrix}<0.
        \end{equation}
        With $V(x) := x^{\mathrm{T}} X x$ and \eqref{eq:P_lmi} we get
        $$
        -V(x^{k+1}) + V(x^{k}) - \gamma^2(u^k)^{\mathrm{T}}(u^k) + (y^k)^{\mathrm{T}} (y^k) < 0.
        $$
        Now we take the sum over $k=0,1,\ldots, T$ which yields
        \begin{equation}\label{eq:dissipativity}
        V(x^0) - V(x^{\mathrm{T}}) + \sum_{k=0}^{T} (y^k)^{\mathrm{T}} (y^k) < \gamma^2 \sum_{k=0}^{\mathrm{T}} (u^k)^{\mathrm{T}}(u^k).
        \end{equation}
        The matrix $X$ is positive definite according to \eqref{eq:linear_constr_lmi} and therefore, stability condition \eqref{eq:l2_gain} follows from \eqref{eq:dissipativity} with $\gamma_0 = (x^0)^{\mathrm{T}} X (x^0)$, which concludes the proof, note that the strict inequality in \eqref{eq:dissipativity} can be replaced by a non strict inequality by adding an $\epsilon I >0$ to $-\gamma^2 I$ in \eqref{eq:dissipativity_0}.
    \end{proof}
\fi
\begin{corollary}
    \label{cor:incremental_l2_bounded}
    If the parameter set $\tilde{\theta}$ satisfies the constraint \eqref{eq:linear_constr_lmi}, the system $\mathcal{S}$ defined in \eqref{eq:system_s} has a finite incremental $\ell_2$-gain $\gamma_{\mathrm{inc}}^2$ and satisfies stability condition \eqref{eq:incremental_l_2_gain}.
\end{corollary}
The proof follows the same arguments as \corollaryref{cor:l2_bounded}, with $\boldsymbol{\Delta}(z^k) = \Delta\psi(z^k)$, $P=P_{\text{slope}}$. Note that we can assume to have the same initial condition for two different sequences which leads to $\Delta x^0 = (x^0)_a - (x^0)_b = 0$.

\subsection{Model Optimization}
In this section, we explain how we initialize the hidden state $x^0$, the initial parameter set $\tilde{\theta}^0$ and how we ensure that the parameters satisfy the constraint \eqref{eq:linear_constr_lmi} during training.

\subsubsection{\texorpdfstring{Initial hidden state $x^0$}{Initial hidden state}}\label{sec:init_hidden_state}
To initialize the hidden state of the linear system $G$, we use a long short-term memory (LSTM) network as suggested in \citet{mohajerin2019multistep}. The initializer LSTM is trained separately from the prediction network on the sequence $\xi^k = \begin{bmatrix}u^k & y^{k-1}\end{bmatrix}^{\mathrm{T}}$, for $k = -T_{\text{init}}, -T_{\text{init}}+1, \ldots, 0$ where we assume to know the previous system state $y^{k-1}$. The initializer LSTM predicts the next system state $\hat{y}^k_{\text{init}}$, the predictor model then takes the final hidden state of the initializer LSTM as the initial state $x^0$.

\subsubsection{\texorpdfstring{Initial parameter $\tilde{\theta}^0$}{Initial parameter}}\label{sec:init_paramter}
Before training the predictor network $\mathcal{S}$, we find an initial parameter set $\tilde{\theta}^0$ that satisfies condition~\eqref{eq:linear_constr_lmi}. This is done by solving a semi-definite program, where the solution is a feasible parameter set~$\tilde{\theta}^0$, as also suggested in \citet{revay2020convex, pauli2022robustness}.
\begin{remark}
    In this work, we use a trivial objective for the semi-definite program to find a feasible solution. For training, we take the feasible solution as initial parameter set and minimize the prediction error (cf.~\sectionref{sec:loss_function_training}).
\end{remark}

\subsubsection{\texorpdfstring{Loss function for predictor $\mathcal{S}$}{Loss function for predictor}}\label{sec:loss_function_training}
Assuming that a feasible initial parameter set $\tilde{\theta}^0$ is found, we optimize the parameters $\tilde{\theta}$ to minimize the mean squared error loss. A barrier function is used as regularization to keep the parameters away from the constraint boundary. Thus our loss function is defined as
\begin{equation}\label{eq:loss}
    \mathcal{L}(y, \hat{y}) = \frac{1}{T}\sum_{k=0}^T \|\hat{y}^k - y^k\|^2 -\nu  \log \operatorname{det}(-M),
\end{equation}
where $M$ is defined as in~\eqref{eq:linear_constr_lmi} and we use the logarithmic barrier function, which is often used in interior-point methods \citep{boyd2004convex}.
During optimization, it can happen that the new parameters do not satisfy condition \eqref{eq:linear_constr_lmi}. Therefore as suggested in \cite{revay2020convex}, we apply a backtracking line search algorithm for $100$ steps. If no feasible parameter set can be found, training stops.
Problem \ref{pb:learn_linear_system_parameters} can now be stated as the following optimization problem
\begin{equation}\label{eq:learning_problem}
    \min_{\tilde{\theta}, X, T} \mathcal{L}(y,\hat{y}) \qquad \text{such that} \qquad M \prec 0,
\end{equation}
where $\hat{y} = \mathcal{S}(u)$ and $(u,y)\in \mathcal{D}$. Note that even though the constraints in \eqref{eq:learning_problem} are convex, the loss function is non-convex with respect to the parameters $\tilde{\theta}$, that directly follows from the recurrent structure of \eqref{eq:system_s}. We assume that the step size of the optimization is small enough such that a violation of condition \eqref{eq:linear_constr_lmi} is noticed if one eigenvalue of $M$ flips its sign. The decay parameter $\nu$ is decreased by a factor of $10$ after $100$ training epochs and initialized with $0.001$.

\section{Numerical Experiment}\label{sec:numerical_experiment}
\subsection{Dataset}
To generate data for system identification, we use a model of a patrol ship with four-degrees-of-freedom \citep{perez20064}. The model is extended to include effects caused by wind \citep{isherwood1972wind} and waves \citep{hasselmann1973measurements}. To navigate the ship, two fixed-pitch rudder propellers are modeled as actuators. Input sequences are generated via an open-loop controller that performs different maneuvers like turns or circles. In addition to the input sequences that are sent to the actuator, we also assume to have access to simulated wind measurements that include the strength and the angle of attack.
The unknown system to be identified has $n_u = 6$ inputs $u^k = \left[\alpha_{\sin}^k~ \alpha_{\cos}^k ~ V_w^k~  n^k~  \delta_{\text{left}}^k~  \delta_{\text{right}}^k\right]^{\mathrm{T}}$, namely \emph{angle of attack} ($\alpha_{\sin}^k$ and $\alpha_{\cos}^k$), \emph{wind speed} ($V_w^k$), \emph{propeller speed} ($n^k$), and \emph{propeller angle} (left $\delta_{\text{left}}^k$ and right $\delta_{\text{right}}^k$). 
The predicted output refers to $n_y=5$ system states $\hat{y}^k = \left[s^k~ v^k~ p^k~ r^k~ \phi^k\right]^{\mathrm{T}}$, the velocity in two dimensions, \emph{surge} ($s^k$) and \emph{sway} ($v^k$), the angular velocity of \emph{roll} ($p^k$) and \emph{yaw} ($r^k$), and the \emph{roll angle} ($\phi^k$).

The dataset $\mathcal{D}$ consists of $96$ hours of routine samples and $29$ hours of OOD samples \citep{darus-2905_2022}. The measurements are sampled every second. The routine samples are split into $60\%$ \emph{training} $\mathcal{D}_{\text{train}}$, $10\%$ \emph{validation} $\mathcal{D}_{\text{val}}$ and $30\%$ \emph{testing} $\mathcal{D}_{\text{test}}$. The OOD samples $\mathcal{D}_{\text{OOD}}$ are only used for evaluation, and consists of a larger range of the propeller speed and more frequent changes in the rudder angle.

\subsection{Models}
As baselines for the predictor model, we use purely learning-based approaches that do not have parametric constraints. For learning the initial hidden state, we use the same LSTM architecture for all models (cf.~\sectionref{sec:init_hidden_state}). We compare our model $\mathcal{S}$ against: (i) \texttt{ltiRNN}, a RNN in LTI structure~\eqref{eq:system_s}, where we use parameters $\tilde{A}, \tilde{B}_1, \tilde{B}_2, C_1, D_{11}, D_{12}, \tilde{C}_2, \tilde{D}_{21}, X, T$ but no constraints, (ii) \texttt{RNN}, a RNN with bias terms \citep{goodfellow2016deep}, and (iii) \texttt{LSTM}, a LSTM that was proposed by \cite{mohajerin2019multistep} for multi-step prediction of quadcopter motion.
We will call the constrained RNN \texttt{cRNN} and add the finite stability gain $\gamma^2$ in the name. We fix the stability gain $\gamma^2$ before training and treat it as a hyperparameter of the model.

\subsubsection{Models training}
The initializer (cf.~\sectionref{sec:init_hidden_state}) is trained on sequences of length $T_{\text{init}}=50$ and $400$ epochs, and we use the same hyperparameters as for the predictor (details on how the hyperparameters are found are provided in \sectionref{sec:evaluation}).
We train the predictor on sequence length $T_{\text{pred}}=50$ and $2000$ epochs. The learning rate is set to $0.0025$ and we use the \emph{Adam}-optimizer \citep{kingma2015adam}. We apply stochastic gradient descent on batches of size $128$ and evaluate the loss function \eqref{eq:loss} on the entire batch of sequences. 
Training and testing is performed on a Nvidia A100 GPU \footnote{Our code is available on \emph{GitHub} \href{https://github.com/AlexandraBaier/deepsysid}{https://github.com/AlexandraBaier/deepsysid}}. 

\subsection{Evaluation}\label{sec:evaluation}
First, we select the best models by hyperparameter optimization on the validation samples $\mathcal{D}_{\text{val}}$. We then evaluate the models on prediction accuracy of the test samples $\mathcal{D}_{\text{test}}$, empirically evaluate their robustness and further test their generalization ability with the OOD samples $\mathcal{D}_{\text{OOD}}$.

For hyperparameter optimization, we use the root mean squared error (RMSE) that is defined as
\begin{equation}\label{eq:rmse}
    \operatorname{RMSE}(\mathcal{D}_{\text{val}}, m) \vcentcolon= \sqrt{\frac{1}{N_{\text{val}}} \sum_{i=1}^{K_{\text{val}}}\sum_{k=0}^{T_{\text{pred}}} ((y_m^k)_i - (\hat{y}_m^k)_i)^2}    \qquad \text{for } m=1, \ldots, n_y.
\end{equation}
We perform grid search on the size of the hidden states and the number of recurrent layers for the \texttt{LSTM} and the \texttt{RNN} model. The $\operatorname{RMSE}$ is calculated for each state separately, we select the model with the lowest mean $\operatorname{RMSE}$ over all states given by
\begin{equation}\label{eq:mean_rmse}
    \overline{\operatorname{RMSE}}(\mathcal{D}_{\text{val}})\vcentcolon=\frac{1}{n_y} \sum_{m=1}^{n_y}\operatorname{RMSE}(\mathcal{D}_{\text{val}},m).
\end{equation}
\ifreport
    The hyperparameters for each model type are shown in \sectionref{apd:grid_search}.
\fi
The best performing models are evaluated for their $\operatorname{RMSE}$ and $\overline{\operatorname{RMSE}}$ on the test samples~$\mathcal{D}_{\text{test}}$.

For robustness evaluation, we first define a perturbed input sequence as $u_{\text{pert}}\vcentcolon=u+\vartheta$ and the corresponding prediction as $\hat{y}_{\text{pert}} = \mathcal{S}(u_{\text{pert}})$. The worst finite stability gain is then defined as
\begin{equation}
    \gamma^2_* \vcentcolon= \max_{u \in \mathcal{D}_\text{val}}\left(\sup_{\vartheta^k \in \mathbb{R}^{n_u}} \frac{\sum_{k=0}^T \|\hat{y}_{\text{pert}}^k \|^2}{\sum_{k=0}^T \|u_{\text{pert}}^k \|^2}\right).
    \label{eq:opt_stab}
\end{equation}
We maximize the stability gain $\gamma^2$ for each input sequence from the validation set $\mathcal{D}_{\text{val}}$. As optimization variable, we use the perturbation sequence $\vartheta$. This empirically evaluates the stability gain of the trained models. Recall that the $\ell_2$-norm is a measure of the energy content of the sequence and therefore the stability gain $\gamma^2$ (cf.~Definition~\ref{def:finite_gain_stability}) refers to the worst possible amplification of the model.

For the evaluation of the incremental stability gain (see Definition~\ref{def:icnremental_stability}) we define the worst incremental stability gain as
\begin{equation}
    \gamma^2_{\text{inc}*} \vcentcolon=\max_{u \in \mathcal{D}_\text{val}} \left( \sup_{\vartheta^k \in \mathbb{R}^{n_u}} \frac{\sum_{k=0}^T \|\hat{y}_{\text{pert}}^k-\hat{y}^k \|^2}{\sum_{k=0}^{\mathrm{T}} \|u_{\text{pert}}^k- u^k\|^2}\right),
    \label{eq:opt_incr_stab}
\end{equation}
where $\hat{y} = \mathcal{S}(u)$ is the prediction of the unperturbed input sequence $u$. For the nonlinear operator $\mathcal{S}$, the incremental stability gain refers to the Lipschitz constant of $\mathcal{S}$ and is considered as a robustness measure of the neural network \citep{fazlyab2019efficient}. A model with a large incremental $\gamma^2_{\text{inc}}$-gain is sensitive to small input changes.
The learning rate for the optimization is set to $0.001$ and we run $2000$ optimization steps for problem \eqref{eq:opt_stab} and $1000$ steps for problem \eqref{eq:opt_incr_stab}.

Lastly, we test the models on OOD samples $\mathcal{D}_{\text{OOD}}$ to evaluate their generalization capacity. Note that for evaluation, the prediction horizon is set to $900$ seconds.

\subsection{Experiment Result}\label{sec:results}
\ifreport
    \begin{table}
        \floatconts
            {tab:evaluatioin_test_set}
            {\caption{Evaluation on test dataset $\mathcal{D}_{\text{test}}$. The best and second-best scores are marked in \textbf{bold} and \textbf{\textit{bold+italic}} respectively.}}
            {\begin{tabular}{rllllll}
	\toprule 
	model & $s$ [m/s] & $v$ [m/s] & $p$ [rad/s] & $r$ [rad/s] & $\phi$ [rad] &   $\overline{\operatorname{RMSE}}(\mathcal{D}_{\text{test}})$ \\ 
	 \midrule 
	 \texttt{cRNN} $\gamma^2 = 5\phantom{0}$ & 0.293 & 0.172 & \textbf{0.00489} & 0.00540 & 0.01820 &  0.09878 \\ 
	 \texttt{cRNN} $\gamma^2 = 10$ & 0.287 & 0.150 &\textbf{0.00489} & 0.00519 & 0.01780 &  0.09306 \\ 
	 \texttt{cRNN} $\gamma^2 = 20$ & 0.292 & 0.138 & \textbf{0.00489} & 0.00525 & 0.01780 &  0.09144 \\ 
	 \texttt{cRNN} $\gamma^2 = 40$ & 0.294 & 0.210 & \textbf{\textit{0.00491}} & 0.00568 & 0.01870 &  0.10660 \\ 
	 \midrule
	 \texttt{ltiRNN} & \textbf{\textit{0.163}} & 0.059 & 0.00511 & 0.00230 & 0.00753 &  0.04724 \\ 
	 \texttt{RNN} & 0.173 & \textbf{\textit{0.055}} & 0.00531 & \textbf{\textit{0.00205}} & \textbf{0.00715} &  \textbf{\textit{0.04845}} \\ 
	 \texttt{LSTM} & \textbf{0.096} & \textbf{0.052} & 0.00590 & \textbf{0.00203} & \textbf{\textit{0.00736}} &  \textbf{0.03259} \\ 
	\bottomrule 
\end{tabular}
}
    \end{table}
\fi
\tableref{tab:evaluatioin_test_set} compares $\operatorname{RMSE}(\mathcal{D}_{\text{test}}, m)$ for $m=1,\ldots,n_y$ of the predicted states for the selected models. The smallest values refer to the highest score according to \eqref{eq:rmse}.
\ifreport     
    \tableref{tab:ood_test} shows the $\operatorname{RMSE}(\mathcal{D}_{\text{OOD}}, m)$ on the OOD samples for all states as well as the $\overline{\operatorname{RMSE}}(\mathcal{D}_{\text{OOD}})$.
    \begin{table}
        \floatconts
            {tab:ood_test}
            {\caption{Evaluation on out-of-distribution data $\mathcal{D}_{\text{OOD}}$. The best and second-best scores are marked in \textbf{bold} and \textbf{\textit{bold+italic}} respectively.}}
            {\begin{tabular}{rllllll}
	\toprule 
	model & $s$ [m/s] & $v$ [m/s] & $p$ [rad/s] & $r$ [rad/s] & $\phi$ [rad] &  $\overline{\operatorname{RMSE}}(\mathcal{D}_{\text{OOD}})$ \\ 
	 \midrule 
	\texttt{cRNN} $\gamma^2 = 5$ & \textbf{\textit{0.520}} & 0.414 & 0.00702 & 0.0137 & 0.0375 &  \textbf{\textit{0.1984}} \\ 
	\texttt{cRNN} $\gamma^2 = 10$ & 0.535 & 0.403 & 0.00692 & 0.0139 & 0.0360 &  0.1991 \\ 
	\texttt{cRNN} $\gamma^2 = 20$ & 0.525 & 0.416 & 0.00697 & 0.0145 & 0.0370 &  0.1999 \\ 
	\texttt{cRNN} $\gamma^2 = 40$ & 0.539 & 0.462 & 0.00729 & 0.0129 & 0.0357 &  0.2114 \\ 
	\midrule
	\texttt{ltiRNN} & \textbf{0.370} & 0.311 & \textbf{0.00599} & \textbf{\textit{0.00713}} & 0.0195 &  \textbf{0.1428} \\ 
	\texttt{RNN} & 0.737 & \textbf{0.261} & 0.00632 & \textbf{0.00573} & \textbf{0.0184} &  0.2057 \\ 
	\texttt{LSTM} & 0.673 & \textbf{\textit{0.293}} & \textbf{\textit{0.00616}} & 0.00581 & \textbf{\textit{0.0188}} &  0.1992 \\ 
	\bottomrule 
\end{tabular}}
    \end{table} 
\fi 
Note that all models have a worse mean RMSE for the OOD samples in comparison with the test set $\mathcal{D}_{\text{test}}$, which is expected since the input range of the propeller speed is larger compared to the test set.
However, the accuracy drop for the constrained RNN is lower than for the learning-based models without constraints.
Despite the higher accuracy of the baseline models (\texttt{LSTM}, \texttt{RNN}, \texttt{ltiRNN}) compared to the constrained RNN (\texttt{cRNN}) no upper bounds on the (incremental) stability gain are available. We show $\gamma^2_*$ and $\gamma^2_{\text{inc}*}$ and relate it to the prediction accuracy on the test samples in \figureref{fig:eval_stability}. The dashed vertical lines refer to the guaranteed upper (incremental) stability bound for the constrained RNN.

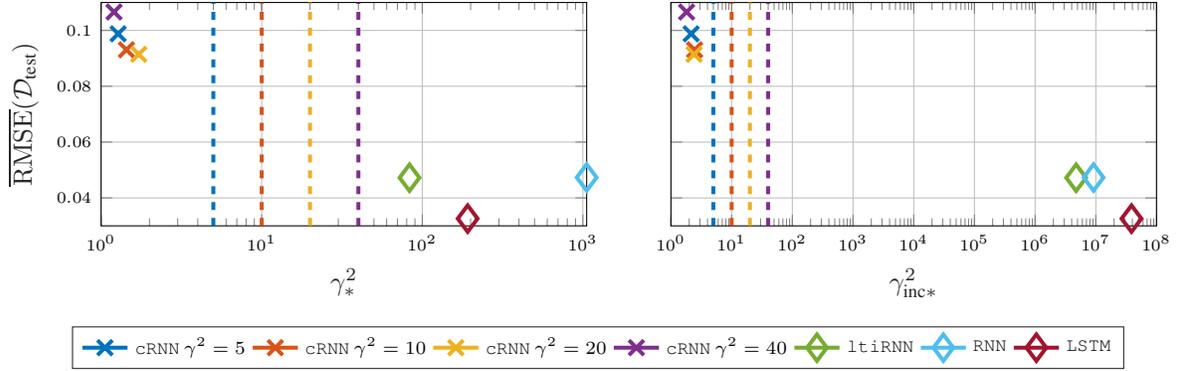
\begin{figure}
    \floatconts
    {fig:eval_stability}
    {\caption{The figure highlights the trade-off between robustness and accuracy when identifying the ship motion for baseline models (\texttt{LSTM}, \texttt{RNN}, \texttt{ltiRNN}) and constrained RNN (\texttt{cRNN}) with different $\gamma^2$ values. On the $y$-axis, the $\overline{\operatorname{RMSE}}$ \eqref{eq:mean_rmse} of the test set is shown. On the $x$-axis, the worst empirical stability gain $\gamma_{*}^2$ (left) and the worst empirical incremental stability gain $\gamma_{\text{inc}*}^2$ (right) are shown.}}
    { 
%
\definecolor{mycolor1}{rgb}{0.00000,0.44700,0.74100}%
\definecolor{mycolor2}{rgb}{0.85000,0.32500,0.09800}%
\definecolor{mycolor3}{rgb}{0.92900,0.69400,0.12500}%
\definecolor{mycolor4}{rgb}{0.49400,0.18400,0.55600}%
\definecolor{mycolor5}{rgb}{0.46600,0.67400,0.18800}%
\definecolor{mycolor6}{rgb}{0.30100,0.74500,0.93300}%
\definecolor{mycolor7}{rgb}{0.63500,0.07800,0.18400}%
\begin{tikzpicture}

\begin{axis}[%
width=0.391\textwidth,
height=0.18\textwidth,
at={(0\textwidth,0\textwidth)},
scale only axis,
xmode=log,
xmin=1,
xmax=1056.0859375,
xminorticks=true,
xlabel style={font=\color{white!15!black}},
xlabel={$\gamma_{*}^2$},
ymin=0.03,
ymax=0.11,
tick label style={/pgf/number format/fixed},
y tick label style={font=\tiny},
x tick label style={font=\tiny},
ylabel style={font=\color{white!15!black}},
ylabel={$\overline{\operatorname{RMSE}}(\mathcal{D}_{\text{test}})$},
axis background/.style={fill=white},
xmajorgrids,
ymajorgrids
]
\addplot [color=mycolor1, line width=1.5pt, mark size=4.0pt, mark=x, mark options={solid, mycolor1}, forget plot]
  table[row sep=crcr]{%
1.27724063396454	0.0987832113393562\\
};
\addplot [color=mycolor2, line width=1.5pt, mark size=4.0pt, mark=x, mark options={solid, mycolor2}, forget plot]
  table[row sep=crcr]{%
1.43829369544983	0.0930614462387725\\
};
\addplot [color=mycolor3, line width=1.5pt, mark size=4.0pt, mark=x, mark options={solid, mycolor3}, forget plot]
  table[row sep=crcr]{%
1.71663296222687	0.091437925199366\\
};
\addplot [color=mycolor4, line width=1.5pt, mark size=4.0pt, mark=x, mark options={solid, mycolor4}, forget plot]
  table[row sep=crcr]{%
1.20344483852386	0.106603728417982\\
};
\addplot [color=mycolor5, line width=1.5pt, mark size=5.0pt, mark=diamond, mark options={solid, mycolor5}, forget plot]
  table[row sep=crcr]{%
83.2603454589844	0.0472353043205477\\
};
\addplot [color=mycolor6, line width=1.5pt, mark size=5.0pt, mark=diamond, mark options={solid, mycolor6}, forget plot]
  table[row sep=crcr]{%
1056.0859375	0.047323550768938\\
};
\addplot [color=mycolor7, line width=1.5pt, mark size=5.0pt, mark=diamond, mark options={solid, mycolor7}, forget plot]
  table[row sep=crcr]{%
191.702728271484	0.0325906837052118\\
};
\addplot [color=mycolor1, dashed, line width=1.5pt, forget plot]
  table[row sep=crcr]{%
5	0.03\\
5	0.0388888888888889\\
5	0.0477777777777778\\
5	0.0566666666666667\\
5	0.0655555555555556\\
5	0.0744444444444445\\
5	0.0833333333333333\\
5	0.0922222222222222\\
5	0.101111111111111\\
5	0.11\\
};
\addplot [color=mycolor2, dashed, line width=1.5pt, forget plot]
  table[row sep=crcr]{%
10	0.03\\
10	0.0388888888888889\\
10	0.0477777777777778\\
10	0.0566666666666667\\
10	0.0655555555555556\\
10	0.0744444444444445\\
10	0.0833333333333333\\
10	0.0922222222222222\\
10	0.101111111111111\\
10	0.11\\
};
\addplot [color=mycolor3, dashed, line width=1.5pt, forget plot]
  table[row sep=crcr]{%
20	0.03\\
20	0.0388888888888889\\
20	0.0477777777777778\\
20	0.0566666666666667\\
20	0.0655555555555556\\
20	0.0744444444444445\\
20	0.0833333333333333\\
20	0.0922222222222222\\
20	0.101111111111111\\
20	0.11\\
};
\addplot [color=mycolor4, dashed, line width=1.5pt, forget plot]
  table[row sep=crcr]{%
40	0.03\\
40	0.0388888888888889\\
40	0.0477777777777778\\
40	0.0566666666666667\\
40	0.0655555555555556\\
40	0.0744444444444445\\
40	0.0833333333333333\\
40	0.0922222222222222\\
40	0.101111111111111\\
40	0.11\\
};
\end{axis}

\begin{axis}[%
width=0.391\textwidth,
height=0.18\textwidth,
at={(0.459\textwidth,0\textwidth)},
scale only axis,
xmode=log,
xmin=1,
xmax=100000000,
xminorticks=true,
xlabel style={font=\color{white!15!black}},
xlabel={$\gamma^2_{\text{inc}*}$},
ymin=0.03,
ymax=0.11,
ytick={0.04,0.06,0.08,0.1},
x tick label style={font=\tiny},
yticklabels={\empty},
axis background/.style={fill=white},
xmajorgrids,
ymajorgrids,
legend style={at={(-0.15,-0.45)}, anchor=north, legend columns=7, legend cell align=left, align=left, draw=white!15!black}
]
\addplot [color=mycolor1, line width=1.5pt, mark size=4.0pt, mark=x, mark options={solid, mycolor1}]
  table[row sep=crcr]{%
2.15228939056396	0.0987832113393562\\
};
\addlegendentry{\scriptsize \texttt{cRNN} $\gamma^2 = 5$}

\addplot [color=mycolor2, line width=1.5pt, mark size=4.0pt, mark=x, mark options={solid, mycolor2}]
  table[row sep=crcr]{%
2.45394277572632	0.0930614462387725\\
};
\addlegendentry{\scriptsize \texttt{cRNN} $\gamma^2 = 10$}

\addplot [color=mycolor3, line width=1.5pt, mark size=4.0pt, mark=x, mark options={solid, mycolor3}]
  table[row sep=crcr]{%
2.40772795677185	0.091437925199366\\
};
\addlegendentry{\scriptsize \texttt{cRNN} $\gamma^2 = 20$}

\addplot [color=mycolor4, line width=1.5pt, mark size=4.0pt, mark=x, mark options={solid, mycolor4}]
  table[row sep=crcr]{%
1.81125950813293	0.106603728417982\\
};
\addlegendentry{\scriptsize \texttt{cRNN} $\gamma^2 = 40$}

\addplot [color=mycolor5, line width=1.5pt, mark size=5.0pt, mark=diamond, mark options={solid, mycolor5}]
  table[row sep=crcr]{%
4739864	0.0472353043205477\\
};
\addlegendentry{\scriptsize \texttt{ltiRNN}}

\addplot [color=mycolor6, line width=1.5pt, mark size=5.0pt, mark=diamond, mark options={solid, mycolor6}]
  table[row sep=crcr]{%
9171487	0.047323550768938\\
};
\addlegendentry{\scriptsize \texttt{RNN}}

\addplot [color=mycolor7, line width=1.5pt, mark size=5.0pt, mark=diamond, mark options={solid, mycolor7}]
  table[row sep=crcr]{%
38713532	0.0325906837052118\\
};
\addlegendentry{\scriptsize \texttt{LSTM}}

\addplot [color=mycolor1, dashed, line width=1.5pt, forget plot]
  table[row sep=crcr]{%
5	0.03\\
5	0.0388888888888889\\
5	0.0477777777777778\\
5	0.0566666666666667\\
5	0.0655555555555556\\
5	0.0744444444444445\\
5	0.0833333333333333\\
5	0.0922222222222222\\
5	0.101111111111111\\
5	0.11\\
};
\addplot [color=mycolor2, dashed, line width=1.5pt, forget plot]
  table[row sep=crcr]{%
10	0.03\\
10	0.0388888888888889\\
10	0.0477777777777778\\
10	0.0566666666666667\\
10	0.0655555555555556\\
10	0.0744444444444445\\
10	0.0833333333333333\\
10	0.0922222222222222\\
10	0.101111111111111\\
10	0.11\\
};
\addplot [color=mycolor3, dashed, line width=1.5pt, forget plot]
  table[row sep=crcr]{%
20	0.03\\
20	0.0388888888888889\\
20	0.0477777777777778\\
20	0.0566666666666667\\
20	0.0655555555555556\\
20	0.0744444444444445\\
20	0.0833333333333333\\
20	0.0922222222222222\\
20	0.101111111111111\\
20	0.11\\
};
\addplot [color=mycolor4, dashed, line width=1.5pt, forget plot]
  table[row sep=crcr]{%
40	0.03\\
40	0.0388888888888889\\
40	0.0477777777777778\\
40	0.0566666666666667\\
40	0.0655555555555556\\
40	0.0744444444444445\\
40	0.0833333333333333\\
40	0.0922222222222222\\
40	0.101111111111111\\
40	0.11\\
};
\end{axis}
\end{tikzpicture}%
}
\end{figure}

\ifreport
    \figureref{fig:eval_trajectories} shows the prediction of the velocities of the best performing constrained RNN (\texttt{cRNN $\gamma^2 = 20$}) \subfigref{fig:crnn_20} and the best performing purely learned model \texttt{LSTM} \subfigref{fig:lstm}.
    \begin{figure}
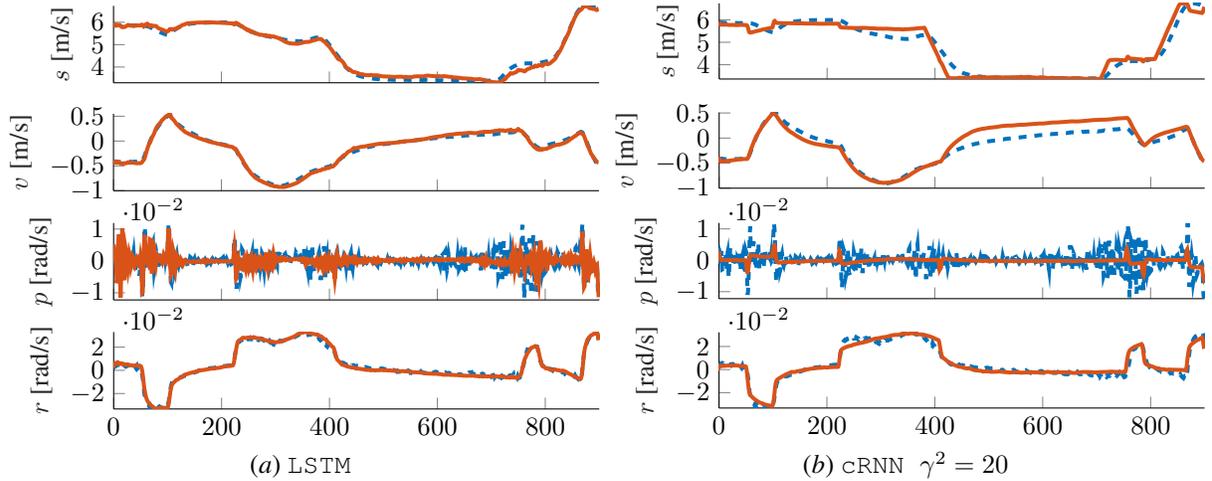

        \floatconts
        {fig:eval_trajectories}
        {\caption{Comparison of different identification methods, the dashed red line refers to the true measurement of the velocities and the solid blue line to the prediction of the model.}}
        {%
        \subfigure[\texttt{LSTM}]{\label{fig:lstm}%
            \input{fig/plots/state_trajectory_LSTM+Init-128-2.tex}}
        \subfigure[\texttt{cRNN $\gamma^2 = 20$}]{\label{fig:crnn_20}%
            \input{fig/plots/state_trajectory_ConstrainedRnn-20000000-64.tex}}
        }
    \end{figure}
    For an intuition on the ship movement \figureref{fig:position} shows the position trajectory for the different models.
    \begin{figure}
        \floatconts
        {fig:position}
        {\caption{Trajectory of ship in 2D plane.}}
        {\input{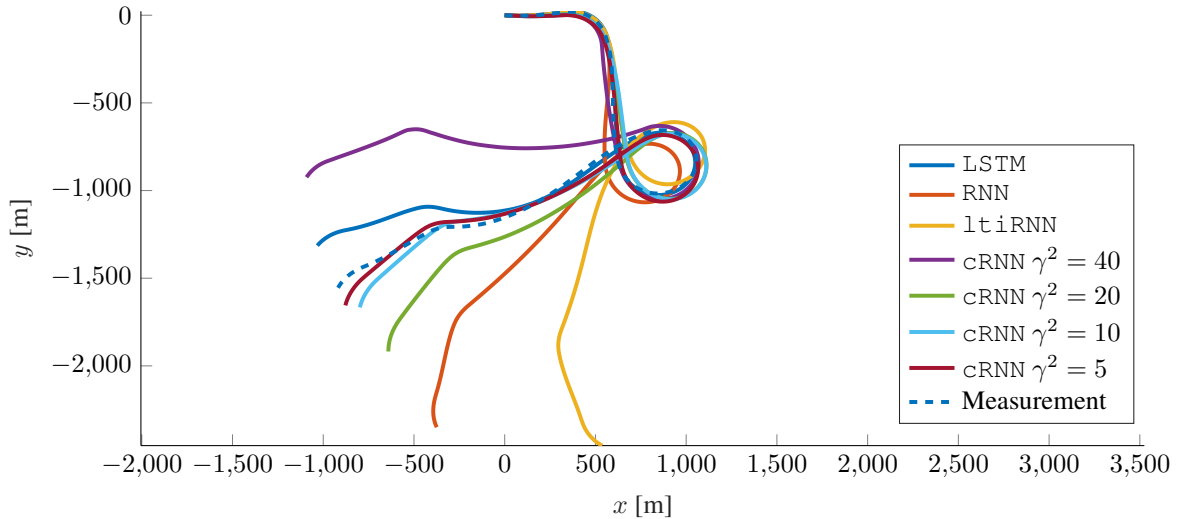}}
    \end{figure}
\fi

\subsection{Discussion}
The trade-off between robustness and accuracy of the predictor model is visualized in \figureref{fig:eval_stability}. Here we can see that for the learned models without constraints, we identify worse (incremental) stability values compared to the guaranteed bounds of the \texttt{cRNN} models. As expected, we thus observe a better generalization of the \texttt{cRNN} for the OOD dataset in comparison to the unconstrained approaches, i.e., the ratio $\overline{\operatorname{RMSE}}(\mathcal{D}_{\text{OOD}}) / \overline{\operatorname{RMSE}}(\mathcal{D}_{\text{test}})$ is smaller for the \texttt{cRNN} models.

Another observation from \figureref{fig:eval_stability} is that the upper bounds are not tight. This means that we were not able to find input sequences that lead to an (incremental) stability gain that was close to the upper bound. Because we use linear functions to bound the non-linearity \eqref{eq:sector_bounded}, \eqref{eq:slope_restriction} and only consider static multipliers $T$ in \eqref{eq:P_lmi}, our constraints are quite conservative.

For the constrained RNN networks, we expect the accuracy to be higher for larger (incremental) stability gains. This hypothesis results from the definition of the allowed parameter set \eqref{eq:linear_constr_lmi} where we have a larger set for higher $\gamma^2$ values. In \tableref{tab:evaluatioin_test_set}, one observes that this hypothesis does not hold in our case. A possible explanation for this behavior is that the initial parameter set $\theta^0$ comes from an optimization problem with trivial objective (cf.~\sectionref{sec:init_paramter}). Since we perform a backtracking line search if parameters are outside the constraint set, training might stop at a local minimum that is close to the constraint boundary. Other initialization methods could improve the prediction results.

We speculate that the roll motion is mostly driven by waves that are induced by wind. As measurement input, we are only able to measure the wind force and leave it to the network to learn an internal wind-to-wave model. The unconstrained models allow for system optimization exploring larger areas of the parameter space compared to the limited region of the constrained RNN and therefore might be able to learn such complex dynamics.

\section{Conclusion}\label{sec:conclusion}
In this work, we used constrained RNNs to identify the dynamics of a ship that moves in open water. With a general framework from robust control, we were able to guarantee both a finite stability gain and a finite incremental stability gain, which serves as additional safety measure for the underlying RNN. We compare the constrained RNN against learning-based approaches without constraints. The evaluation shows that there exists a trade-off between robustness and prediction accuracy.
We empirically evaluated the stability of the models by finding input sequences that lead to large (incremental) stability gains. These sequences refer to potentially dangerous predictions of the model like swing-up of the roll angle. We further showed that robust RNNs slightly generalize better to unseen input sequences (OOD dataset) compared to LSTMs and RNNs.


\acks{We thank Alexandra Baier for helpful discussions and her python package \emph{deepsysid} that was used for training, testing, and evaluating our models. This work is funded by Deutsche Forschungsgemeinschaft (DFG, German Research Foundation) under Germany's Excellence Strategy - EXC 2075 – 390740016. We acknowledge the support by the Stuttgart Center for Simulation Science (SimTech). The authors thank the International Max Planck Research School for Intelligent Systems (IMPRS-IS)
for supporting Daniel Frank.}

\bibliography{bib}

\ifreport
	\appendix

	\section{Hyperparameter Optimization}\label{apd:grid_search}
	
\ifreport
    Hyperparmeter optimization for different model types. The size of the hidden state for \texttt{LSTM} and \texttt{RNN} is denoted by $n_h$ and the size of $z^k = w^k$ for the linear model $G$ \eqref{eq:system_s} is denoted by $n_w$. $\overline{\operatorname{RMSE}(\mathcal{D}_{\text{val}})}$ refers to the mean RMSE of the validation set (cf. \eqref{eq:rmse}), the models are listed in ascending order.
    \begin{longtable}{rlllllllll}
	\toprule 
	Model  & $\overline{\operatorname{RMSE}(\mathcal{D}_{\text{val}})}$ & $\boldsymbol{\gamma}^2$ & $n_w$ & $n_h$ & \# rec. & \# prediction & $t_{\text{train}}$ & final barrier & max gradient \\ 
	type & & & & & layers & steps & (hh:mm:ss) & value & norm \\ 
	 \midrule 
	 \texttt{cRNN} & \textbf{0.09223} & 20 & 64 & - & - & 427 & 03:10:36 & -0.001 & 1.8 \\ 
	 \texttt{cRNN} & 0.09282 & 20 & 128 & - & - & 349 & 02:48:13 & -0.013 & 1.8  \\ 
	 \texttt{cRNN} & 0.09351 & 20 & 256 & - & - & 521 & 04:18:14 & -0.00019 & 1.8  \\ 
	 \texttt{cRNN} & 0.09359 & 20 & 192 & - & - & 417 & 03:47:26 & -0.0016 & 1.8  \\ 
	 \texttt{cRNN} & 0.09363 & 10 & 64 & - & - & 511 & 03:52:36 & -7.3e-05 & 2.1  \\ 
	 \texttt{cRNN} & 0.0956 & 10 & 128 & - & - & 336 & 03:03:06 & -0.0095 & 2.1  \\ 
	 \texttt{cRNN} & 0.09584 & 10 & 192 & - & - & 503 & 04:23:39 & -0.00011 & 2  \\ 
	 \texttt{cRNN} & 0.09734 & 10 & 256 & - & - & 526 & 04:20:45 & -0.00014 & 2.4  \\ 
	 \texttt{cRNN} & 0.09822 & 5 & 64 & - & - & 344 & 01:51:38 & -0.0052 & 2.7  \\ 
	 \texttt{cRNN} & 0.102 & 5 & 256 & - & - & 522 & 04:20:24 & -0.0001 & 2.7  \\ 
	 \texttt{cRNN} & 0.102 & 5 & 128 & - & - & 446 & 03:03:06 & -0.00069 & 2.7  \\ 
	 \texttt{cRNN} & 0.1059 & 5 & 192 & - & - & 512 & 04:22:33 & -8.7e-05 & 2.4  \\ 
	 \texttt{cRNN} & 0.1061 & 40 & 64 & - & - & 503 & 04:03:41 & -0.00013 & 2.3  \\ 
	 \texttt{cRNN} & 0.1075 & 40 & 128 & - & - & 523 & 04:45:00 & -0.00017 & 2.3  \\ 
	 \texttt{cRNN} & 0.1078 & 40 & 192 & - & - & 758 & 04:52:58 & -2.1e-06 & 2.3  \\ 
	 \texttt{cRNN} & 0.1078 & 40 & 256 & - & - & 525 & 04:39:11 & -0.00025 & 2.3  \\ 
	 \midrule 
	 \texttt{LSTM} & \textbf{0.03181} & - & - & 128 & 2 & 2000 & 00:28:16 & - & - \\ 
	 \texttt{LSTM} & 0.03605 & - & - & 192 & 2 & 2000 & 00:27:44 & - & -  \\ 
	 \texttt{LSTM} & 0.03725 & - & - & 128 & 3 & 2000 & 00:27:19 & - & -  \\ 
	 \texttt{LSTM} & 0.0388 & - & - & 128 & 4 & 2000 & 00:26:31 & - & -  \\ 
	 \texttt{LSTM} & 0.04076 & - & - & 192 & 3 & 2000 & 00:30:21 & - & -  \\ 
	 \texttt{LSTM} & 0.0409 & - & - & 64 & 3 & 2000 & 00:28:32 & - & -  \\ 
	 \texttt{LSTM} & 0.04121 & - & - & 64 & 2 & 2000 & 00:28:15 & - & -  \\ 
	 \texttt{LSTM} & 0.04369 & - & - & 256 & 2 & 2000 & 00:28:39 & - & -  \\ 
	 \texttt{LSTM} & 0.04406 & - & - & 192 & 4 & 2000 & 00:33:51 & - & -  \\ 
	 \texttt{LSTM} & 0.04768 & - & - & 256 & 3 & 2000 & 00:28:19 & - & -  \\ 
	 \texttt{LSTM} & 0.05025 & - & - & 64 & 4 & 2000 & 00:26:06 & - & -  \\ 
	 \texttt{LSTM} & 0.05125 & - & - & 256 & 4 & 2000 & 00:29:05 & - & -  \\ 
	 \midrule 
	 \texttt{RNN} & \textbf{0.04732} & - & - & 64 & 2 & 2000 & 00:31:39 & - & - \\ 
	 \texttt{RNN} & 0.04831 & - & - & 64 & 3 & 2000 & 00:26:14 & - & -  \\ 
	 \texttt{RNN} & 0.04869 & - & - & 64 & 4 & 2000 & 00:28:50 & - & -  \\ 
	 \texttt{RNN} & 0.07736 & - & - & 128 & 2 & 2000 & 00:18:23 & - & -  \\ 
	 \texttt{RNN} & 0.08317 & - & - & 192 & 2 & 2000 & 00:40:30 & - & -  \\ 
	 \texttt{RNN} & 0.1065 & - & - & 128 & 3 & 2000 & 00:19:40 & - & -  \\ 
	 \texttt{RNN} & 0.1066 & - & - & 128 & 4 & 2000 & 00:21:16 & - & -  \\ 
	 \texttt{RNN} & 0.11 & - & - & 192 & 4 & 2000 & 00:27:09 & - & -  \\ 
	 \texttt{RNN} & 0.1316 & - & - & 192 & 3 & 2000 & 00:39:56 & - & -  \\ 
	 \texttt{RNN} & 0.133 & - & - & 256 & 4 & 2000 & 00:37:54 & - & -  \\ 
	 \texttt{RNN} & 0.1344 & - & - & 256 & 2 & 2000 & 00:39:59 & - & -  \\ 
	 \texttt{RNN} & 0.1365 & - & - & 256 & 3 & 2000 & 00:16:58 & - & -  \\ 
	 \midrule 
	 \texttt{ltiRNN} & \textbf{0.04512} & - & 192 & - & - & 2000 & 01:23:45 & - & 11 \\ 
	 \texttt{ltiRNN} & 0.04559 & - & 256 & - & - & 2000 & 01:28:58 & - & 12  \\ 
	 \texttt{ltiRNN} & 0.04932 & - & 128 & - & - & 2000 & 01:24:11 & - & 9.4  \\ 
	 \texttt{ltiRNN} & 0.04938 & - & 64 & - & - & 2000 & 01:22:20 & - & 11  \\ 
	\bottomrule 
\end{longtable}

\fi
\fi

\end{document}